\pdfoutput=1
\documentclass[11pt]{article}

\usepackage[]{acl}
\usepackage{multirow}
\usepackage{colortbl}
\usepackage{listings}
\usepackage{times}
\usepackage{latexsym}
\usepackage{graphicx}
\usepackage[T1]{fontenc}
\usepackage[utf8]{inputenc}

\usepackage{microtype}

\usepackage{inconsolata}

\usepackage{listings}
\usepackage{enumitem}

\title{Fighting crime with Transformers: Empirical analysis of address parsing methods in payment data}

\author{Haitham Hammami \\
  \texttt{haitham.hammami\textsuperscript{1}} \\\And
  Louis Baligand \\
  \texttt{louis.baligand\textsuperscript{1}} \\
  \\ \textsuperscript{1}\texttt{@alumni.epfl.ch} \\\And
  Bojan Petrovski \\
  \texttt{bojan.petrovski\textsuperscript{1}}
}

\begin{document}
\maketitle
\begin{abstract}
In the financial industry, identifying the location of parties involved in payments is a major challenge in the context of various regulatory requirements. For this purpose address parsing entails extracting fields such as street, postal code, or country from free text message attributes. While payment processing platforms are updating their standards with more structured formats such as SWIFT with ISO 20022\footnote{\url{https://www.iso20022.org/iso-20022-message-definitions}}, address parsing remains essential for a considerable volume of messages. With the emergence of Transformers and Generative Large Language Models (LLM), we explore the performance of state-of-the-art solutions given the constraint of processing a vast amount of daily data. This paper also aims to show the need for training robust models capable of dealing with real-world noisy transactional data. Our results suggest that a well fine-tuned Transformer model using early-stopping significantly outperforms other approaches. Nevertheless, generative LLMs demonstrate strong zero-shot performance and warrant further investigations.  

%We open-source our fine-tuned models, implementation, and our 700MB dataset augmented on real-world payments.%
\end{abstract}

\section{Introduction}

To ensure adherence with regulatory requirements, it is essential for financial institutions to understand precisely where the money is originating and where it is flowing. The new standard of international payment messages ISO 20022 for SWIFT has the potential to simplify the task of locating payment parties by enabling the beneficiary and originator address to be delivered in a structured format. However, a considerable amount of messages are still delivered with an address in free text form. This problem is further exacerbated by the use of legacy payment processing platforms. Thus \textit{Address Parsing} is required to extract address fields such as street, postal code, city, or country. 

Our work has three main contributions. Firstly, it offers an open-sourced, augmented dataset, addressing the limitations of bench-marking on clean datasets and enabling research on noisy real-world payment data.
Secondly, by empirically analyzing and comparing various techniques, this paper uncovers an effective approach for multinational address parsing on distorted data. Lastly, we open-source the fine-tuned state-of-the-art model, aiding future research and application in a multinational setup written in Latin alphabet and transliterated in ASCII format.
%e.g. English, Spanish, French, German, Czech, Danish, Irish, Italian, Dutch, Norwegian, Polish, Portuguese, Romanian, Slovenian, Swedish and Turkish.

%By empirically analyzing and comparing various state-of-the-art techniques, including but not limited to LibPostal, DeepParse, and Transformer-based models, this paper aims to uncover the most effective strategies for extracting and interpreting address data.%
%There exist off-the-shelf solutions such as \textit{libpostal}\footnote{\url{https://github.com/openvenues/libpostal}} however it does not perform well on noisy data with missing address parts.

%The recent emerging solutions in Natural Language Processing tasks such as Question Answering, Text generation, or Token classification are pushing the boundaries of many hands-on applications and industries, banking included. It has been used for Multinational Address Parsing \citet{yassine2021multinational} and hence is promising in our financial use case.

\section{Related Work}
Looking through prior art, we identified various solutions to the problem at hand. Early rule-based attempts \cite{xu2012geocoding} were shown to be incapable of dealing with the complexities of noise in real-world data.  Generative approaches based on Hidden Markov Models \citep{Li2014HMMbasedAP} and probabilistic Conditional Random Fields (CRF) \citep{CRF} give the first promising results on real-world data. A notable and popular off-the-shelf solution based on the CRF approach is LibPostal, a C library that enables multilingual address parsing and normalization\footnote{\url{https://github.com/openvenues/libpostal}}. As it was trained on millions of multilingual, international addresses, this model shows great potential for robustness when faced with unknown addresses from different countries. Therefore we consider it as the baseline solution used for our benchmarks. 

In recent years, advancements in natural language processing have been driven by deep learning techniques \citep{DBLP:journals/corr/abs-1103-0398}. This is evidenced by solutions like DeepParse \citep{beauchemin2023deepparse}, which leverages a Seq2Seq bidirectional-LSTM neural network architecture, and Transformer-based approaches such as the one proposed by \citet{georeoberta}. The efficacy of Transformer models has also been explored in domain-specific scenarios, as demonstrated in studies performed by \citet{Sahay2023} and \citet{kulkarni-etal-2023-domain}. While these existing frameworks provide valuable insights most of them have either a domain, country, or robustness limitation. Our work seeks to benchmark some of these techniques on payment data and build on them to develop a solution 
tailored to our task while also exploring more innovative approaches.

%\textbf{The disruptive potential of generative AI has been vividly showcased in its recent breakthroughs. Although its most notable use cases are more related to text generation, it was compelling to adapt it to our task. }

%More recent solutions are based on deep learning  like Deepparse introduced by \citetp{beauchemin2023deepparse}, which employs a Seq2Seq bidirectional-LSTM neural network architecture and more recent approach based on Transformers \citep{georeoberta}. Transformer approach is also explored in domain specific scenarios \citep{Sahay2023} \citep{kulkarni-etal-2023-domain}. These existing frameworks provide valuable insights, but our proposed approach aims to build upon and surpass their capabilities through a more cutting-edge and specialized solution.%

%To evaluate performance of Address Parsing models and more generally in Name Entity Recognition, a common approach is to use F1 score \citep{tjong-kim-sang-de-meulder-2003-introduction}. More complex approaches have been explored by \citet{fu-etal-2020-interpretable} introducing various attributes such as Out-of-Vocabulary (OOA) Density or entity length to improve interpretation from models. For simplicity purposes we use F1 score however we customize it by considering OOA.%

\section{Data}

% To produce a PDF file, pdf\LaTeX{} is strongly recommended (over original \LaTeX{} plus dvips+ps2pdf or dvipdf). Xe\LaTeX{} also produces PDF files, and is especially suitable for text in non-Latin scripts.
The basis of our data comes from the training data of 
\textit{DeepParse}\footnote{\url{https://github.com/GRAAL-Research/deepparse-address-data}}
\citep{Yassine_2020}, which is a set of postal addresses across multiple countries, with the exact number of samples per country being reported in their work. This data has been generated using data from the \textit{lipbostal} project, which in itself was trained on real-world addresses from OpenStreetMap. 
Each sample is provided as a tuple: the address (string) and a list of tags, one for each word in the address. A tag can be either StreetName, StreetNumber, Unit, PostalCode, Municipality, or Province. The dataset is comprised of two types of data: \textbf{clean} that have all the aforementioned tags in every sample, except for Unit and PostalCode being optional, and \textbf{incomplete} missing at least one of the other tags. We only sample from the clean data, from which we present an example in Table~\ref{tab:data sample}. It is important to highlight that there is no indication of the country. This is a serious limitation as the main goal of the address parsing in transactional data is country derivation for regulatory purposes. Another limitation is the absence of the beneficiary name, an equally important part of SWIFT messages. These limitations are further discussed in section \ref{data_aug}.

\begin{table*}
\centering
\begin{tabular}{lllll}
\hline
\textbf{StreetName} & \textbf{StreetNumber} & \textbf{PostalCode} & \textbf{Municipality} & \textbf{Province}\\
\hline
jakob-sturm-w. & 35 &  80995 &  munich & bavaria\\

\hline
\end{tabular}
 \caption{Data Sample}
 \label{tab:data sample}
\end{table*}

From the clean data of the original dataset, we extract at most 100K samples from each country's training and testing set and we union the extracted samples to get a total of 3.4M addresses. During sampling, we remove all addresses that are not written in the Latin alphabet as this is the standard in SWIFT messages. Out of these addresses, we resample 100K rows for testing and leave the rest for training as we need a large volume of data to train the large models. Then from a set of addresses from countries that were not used in training and testing sets, we create a \textbf{zero-shot} dataset by extracting at most 100K samples from each country.

From this point onward, we will refer to these datasets as \textbf{synthetic} data, from which we create three versions $V_0$, $V_1$ and $V_2$ as described in section \ref{data_aug}.
\subsection{Augmented Data}
\label{data_aug}
The goal of data augmentation is to mimic as closely as possible the structure of our real-life data, referred to as \textbf{production} data. We performed a comprehensive analysis on production samples comprised of 1,600 manually labeled messages, handpicked in a way that covers the most common addresses from all the countries where data is available. This analysis enabled us to deduce the underlying distribution of the address structure and how straightforward it is to parse it. Since SWIFT messages are in free text, postal address standards are sometimes not respected, as some parts of the address are omitted, misspelled, or wrongfully positioned. Moreover, we can find parts of the message that are completely unrelated to the address itself, such as adding a phone number, or an account number or going to the extent of writing "Address information is in line 4". Based on these observations, we augment our synthetic data as follows:

\begin{enumerate}[label=\Alph*]
\item \textbf{Name}: For adding the name we used the python library \textit{Faker}\footnote{\url{https://faker.readthedocs.io/en/master/}} to generate fake person and company names, which we prepend to the address by adding the tag Name in the corresponding position of the tag list.
\item \textbf{Country}: By analyzing the SWIFT messages at our disposal, we observed that the presence of the country in the address can be in either by displaying its name (in different languages) or its ISO code. By simulating the distribution of these occurrences from our production data, we insert the country in its position in the address with the tag Country or CountryCode. The languages used are English, French, German, Italian, and the country's original language.
\item \textbf{Address structure}: As it was well-examined in \citet{Yassine_2020}, the synthetic data is well-structured, with each address following its country's standard address format, as opposed to production data, where the free-text message may or may not follow these standards. To mitigate this discrepancy, for each address of the synthetic data we randomly sample a production address and we apply its ordering of tags as a mask. This leads to potential changes in the synthetic address, including the removal, addition, or rearrangement of elements. Consequently, some addresses may deviate from the standard structure of the country's address format. This variation aims to prevent the models from overfitting to country-specific address structures, addressing a concern highlighted by \citet{yassine2021multinational}. In their study, they observed a significant drop in model performance on zero-shot data, particularly on addresses from countries with distinct address formats compared to those present in the training data. 
\item \textbf{Line separation}: In around a third of the SWIFT messages we observe that the name and the address are separated by a line return. This is additional information that the model can leverage to perform better parsing, therefore we add to the third of the synthetic data the symbol "\$" between the name and address and give it the tag "HardSep".
\item \textbf{Labeling augmentation:} We adapt our labels to the \textbf{BIO} tagging schema \citep{sang2000introduction} by adding the prefix "B-" to the beginning of the class, and "I-" for words inside of the class. Since the synthetic data is purely address-related and does not have terms outside of the previously stated tags, no word would be tagged as "O". To adapt to our production data, we introduce a new tag: \textbf{OOA} (Out-Of-Address), that is any term found in the SWIFT message unrelated to the address itself. Therefore, while rearranging the address structure, whenever the production sample has an OOA tag, we generate an OOA term and insert it into the synthetic address. This term can be a random number, an alphanumeric code, a postbox number, or a duplicate term from the address itself. The latter is labeled as OOA and not its real tag because it is redundant and often miss-placed, as seen in Table~\ref{tab:Data aug} where the term for OOA was chosen to be a redundant mention of the country. The decision of the type of the generated term is based on the analyzed probability distribution of the nature of this OOA in production data.
\end{enumerate}

    %  we refer as any term found in the SWIFT message that is unrelated to the address itself, as discussed previously, it could be a phone number for example.

During the process of data augmentation, we retain three versions of datasets:
\begin{itemize}
    \item \textbf{$V_0$}:
    the original sampled addresses with no augmentation.
    \item \textbf{$V_1$}: Given $V_0$, we apply basic cleaning and address structure masking technique (rearrangement or removal, but no addition of augmented address parts).
    \item \textbf{$V_2$}: Given $V_0$, we apply all the aforementioned augmentation techniques described in points A to E.
\end{itemize}

\begin{table*}
    \centering
    \begin{tabular}{cc}
         \hline
         \textbf{Synthetic address}& jakob-sturm-w. 35 80995 munich bavaria\\
         \hline
         \textbf{Synthetic tags}& [StreetName, StreetNumber, PostalCode, Municipality, Province]\\
         \hline
         \textbf{Prod mask}& [Name, StreetName, StreetNumber, Municipality, PostalCode, Country, OOA]\\
         \hline
         \textbf{augmented address}& John Doe, jakob-sturm-w. 35 munich 80995 germany germany\\
         \hline
    \end{tabular}
    \caption{Data augmentation example}
    \label{tab:Data aug}
\end{table*}
\section{Proposed Approaches}

Given a free text as string, the task is to parse the address fields. The possible fields should be chosen among "Name", "StreetName", "StreetNumber", "Unit", "Municipality", "PostalCode", "Province" and "Country". Note that not all words need to be parsed.\\

\texttt{Input}: \\
\texttt{"jakob-sturm-w. 35 80995 munich bavaria"} \\

\texttt{Output:}
\begin{verbatim}
    {"StreetName": "jakob-sturm-w.", 
    "StreetNumber": "35", 
    "PostalCode": "80995", 
    "Municipality": "munich",
    "Province": "bavaria"}
\end{verbatim}

\subsection{LibPostal}
Currently being a very prominent tool for our task, it is important to benchmark this model's performance against our subsequent approaches. Since it is using a different labeling schema, we attempt to create a mapping to the set of tags we are using to align it with our metrics. The mapping is provided in the appendix in Table~\ref{tab:postal-mapping}.

\subsection{DeepParse}
Given that we are working with the same training dataset, it is meaningful to use this model as a baseline. For $V_0$ and $V_1$, we are using the same set of tags, therefore we run the inference using the standard model with \textit{Byte-pairs embedding (bpemb)}. Since we are adding more tags into $V_2$, namely \textit{Name}, \textit{Hardsep}, \textit{OOA}, \textit{Country} and \textit{CountryCode}, we retrain the model on the new set of tags, using $V_2$ as training data and the default training parameters proposed in the library. 

\subsection{Transformers}
Given that we are dealing with a token classification task, our first intuition was to experiment with the transformer models family, given their proven track record in performing well with similar tasks, namely for Named Entity Recognition (NER) \citep{liu2021nerbert}. For all the subsequent trials, the experimental setup and hyperparameters are reported in the appendix \ref{sec:setup}.
\subsubsection{Baseline models and initial results}
\label{baseline models}
We started by looking at the distilled versions of the BERT family given their small size and short training duration \cite{sanh2020distilbert}, namely \textit{distilbert-base-uncased}, \textit{distilbert-base-multilingual-cased}, \textit{distilbert-cased} and \textit{distilroberta-base}. The performance of the four transformer models is reported in Table \ref{tab: results table}. The numbers show that the model distilbert outperforms the other models in every set, while also being the most robust model across its copies (folds) since it has the smallest standard deviation, which makes us confident to perform our subsequent experimentation using this model.
\subsubsection{Further experiments}
%While the task is defined as Token Classification, where each word has its own tag, the nature of our data suggests another approach to redefining this task. Each address can be segmented into contiguous blocks, with each tag assigned to at most one block. We wanted to test the hypothesis of whether enforcing this constraint onto the model would make it perform better. Therefore we implemented two different custom losses to use for model training:
%\begin{itemize}
    %\item \textbf{CustomLoss1}: If each tag appears in at most one block, then any tag of the form \textit{B-tag} appears at most one, so the idea is to penalize its appearance more than once. If we set $N_Bi$ to be the number of appearances of the B-tag \textit{i} in the sentence, then the penalty translates to:
    %\[ P_1 = \sum_{i}^{N} \max(0, \text{{N_Bi}} - 1) \]

    %\item \textbf{CustomLoss2}: If each block has at most one tag, then any \textit{I-tag} can only be preceded by the same tag (B or I)
    
\begin{figure}[h]
  \centering
  \includegraphics[width=0.5\textwidth]{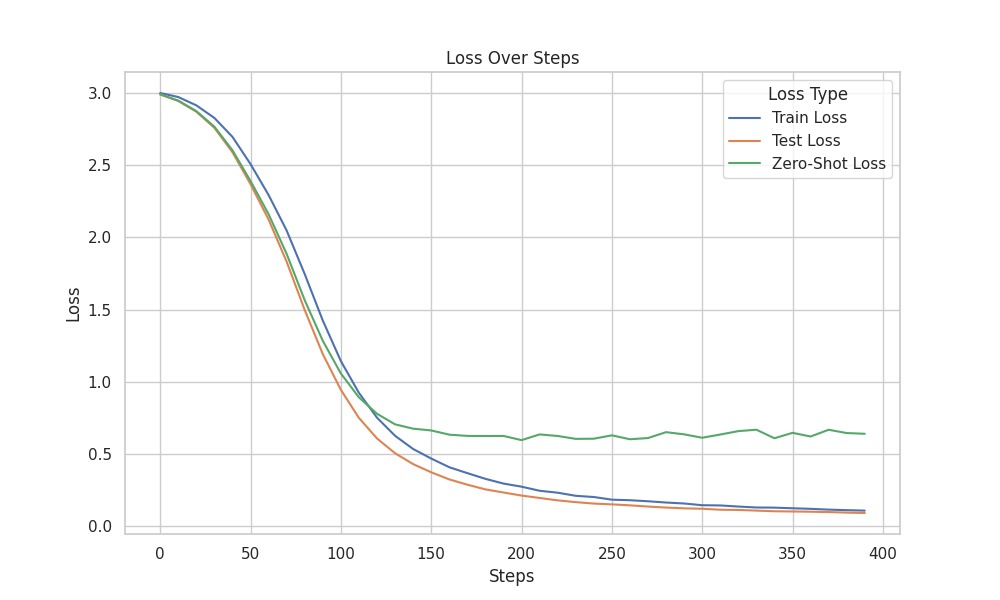}
  \caption{Learning curve showing the model's overfit}
  \label{fig:learning_curve}
\end{figure}

%\end{itemize}
We attempt to challenge our early assumptions on the way we augment and generate the data by modifying $V_2$ in two separate ways: removing the prefix from the tag and removing the HardSep token. Moreover, Figure~\ref{fig:learning_curve} suggests overfitting, prompting us to experiment with early stopping, using cross-entropy loss on zero-shot data as the stopping metric.
For completeness, we retrain the model using $V_0$ and $V_1$ training datasets to have comprehensive bench-marking with LibPostal and DeepParse.

The results reported in \ref{Results}, further discussed later, show that the only variation that substantially improved the model's performance is the addition of early stopping. Using the best performing configuration, we switch to larger transformer models, namely \textit{bert-base-uncased}, \textit{bert-large-uncased} and \textit{xlm-roberta-large}.

\subsection{Decoder Based Approach}

Given the various legal and regulatory constraints stemming from bank privacy laws especially when dealing with cross-jurisdictional payments, API-based LLMs such as GPT-4 are not something we could productionalize. Furthermore, as we are dealing with large volumes of daily data our focus was on "small" sized LLMs that have the best performance-to-cost ratio. 

We chose Llama 2 7b \citep{touvron2023llama} and its derivative Mistral-7B \citep{jiang2023mistral} as benchmark models. We designed several prompts following Prompt Engineering Best Practices\footnote{https://www.promptingguide.ai/introduction/tips}\footnote{https://help.openai.com/en/articles/6654000-best-practices-for-prompt-engineering-with-openai-api} and selected the best performing one. Our full prompt can be found in the appendix in Figure~\ref{fig:prompt}. To tackle cases where the same word appears twice in the input as two different tags, we prefix each word with its index. This will enable us to easily post-process the output in the desired structured format. Given that we are using "small" sized LLMs we did not need to quantize the models. 

We use 1,000 addresses from our $V_2$ zero-shot dataset to benchmark both LLAMA 2 7b (Chat) and Mistral-7B (Instruct). As Mistral-7B performed substantially better our further trials focused on extracting more performance from it by experimenting with the inference sampling parameters and by attempting to train the LORA adapter \citep{DBLP:journals/corr/abs-2106-09685} for our specific task.

%In gerran Among the wide range of promising LLMs, we run a first set of experiments using standard Llama 2 7B Chat Huggingface version . Next,  Mistral 7B Instruct v0.2  and Mixtral 8x7B Intruct v0.1. We first run with single prompt inference to benchmark the standard performance using a temperature=0.2 and min\_p=0.1. We then use 1,000 samples from the training dataset to do Supervised Fine-Tuning (SFT) with 3 epochs. We arbitrarily design a prompt following Prompt Engineering Best Practices that can be found in the appendix in Figure~\ref{fig:prompt}. To tackle cases where the same word appears twice in the input, we concatenate each word with its index. For standard models, we keep the default floating point precision. To manage costs and time constraints, we limit our training and testing dataset to 10,000 samples. We stop the inference after seeing two consecutive jump lines and we parse the output in the format of JSON.%

\subsection{Metrics}
We compute the Precision and Recall score based on each tag, we then average the F1 score overall and also show the standard deviation based on each of the transformer-based trained versions for each model and each dataset. To keep the purpose of the application general we use F1 score without giving more weight to precision or recall.
For LLMs, we use a 10,000 random sample from the training and zero-shot dataset. There is no standard deviation in this case.

\section{Results and discussion}
\label{Results}

\subsection{Transformer Based Approach}
We summarize our findings in Table~\ref{tab: results table}. The main takeaways from it are:

     \textbf{- train/test scores}: what stands out at first glance is how small the difference between these scores is across all experiments. This could indicate how well the models can perform the task. However, this resemblance makes it hard to spot over-fitting, which makes us focus on the zero-shot score as the basis of our comparison. \\
     
     \textbf{- LibPostal:} The inference scores for the three datasets are very consistent, and dropping across the data versions. This is an expected result as LibPostal had a large multinational training set so it was equally exposed to all countries, but its training data highly resembles $V_0$, so its performance drops as the data gets more noisy. The main observed limitation of LibPostal comes from the fact that it only expects address-related parts and complete addresses, which is seldom the case with payment data and consequently the $V_2$ dataset. \\
     
     \textbf{- DeepParse:} The results of $V_0$ matches the reported numbers of \citet{Yassine_2020}. What's interesting is the improvement in the zero-shot score between $V_0$ and $V_1$, also observed for distilbert. This is caused by the alteration of the address structure that produced a form of \textit{Data Leakage} that is necessary for robustness but should be nonetheless taken into consideration when comparing the models \cite{elangovan2021memorization}. For $V_2$, the retrained model failed to produce satisfactory results. \\
     
     \textbf{- DistilBERT:} DistilBERT surpassed DeepParse on all data versions, including outperforming Postal on $V_1$ and $V_2$. This underscores the transformer models' adaptability and strong generalization capability across unseen data during training. Specifically, in our comparison among transformers (as discussed in \ref{baseline models}), the DistilBERT uncased version yielded the best results, aligning with the uncased nature of our data; this is further supported by the lower scores of the cased version. \\Continuing our experimentation with the uncased model, we see that removing the \textit{HardSep} token does not significantly improve the performance, so we decide to keep it as we deem that it will be even more useful on more noisy production data. Removing the prefix yields a slightly better score, but a more notable improvement comes with the early stopping. Combining the last two modifications produces a lower score than just having early stopping so we keep it as the sole modification on our training pipeline. \\
     
     \textbf{- Larger models:} We observe a clear pattern of the score increasing as we increase the size of the model, with the best result achieved by XLM-RoBERTa-Large, despite being a cased model. Testing this model against a few other selected models in predicting production data, we see that it outperforms them still, with the observation of a clear correlation between zero-shot and prod scores, which confirms the justification of using the former as the basis of our comparison.\\

\begin{table*}[]
\begin{tabular}{
>{\columncolor[HTML]{FFFFFF}}c |
>{\columncolor[HTML]{FFFFFF}}c |
>{\columncolor[HTML]{FFFFFF}}c 
>{\columncolor[HTML]{FFFFFF}}c |
>{\columncolor[HTML]{FFFFFF}}c 
>{\columncolor[HTML]{FFFFFF}}c |
>{\columncolor[HTML]{FFFFFF}}c 
>{\columncolor[HTML]{FFFFFF}}c |
>{\columncolor[HTML]{FFFFFF}}c }
\cellcolor[HTML]{FFFFFF}                                                                                   & \cellcolor[HTML]{FFFFFF}                                         & \multicolumn{2}{c}{\cellcolor[HTML]{FFFFFF}\textbf{train}} & \multicolumn{2}{c}{\cellcolor[HTML]{FFFFFF}\textbf{test}} & \multicolumn{2}{c}{\cellcolor[HTML]{FFFFFF}\textbf{zero-shot}} \\
\multirow{-2}{*}{\cellcolor[HTML]{FFFFFF}\textbf{\begin{tabular}[c]{@{}c@{}}Data \\ Version\end{tabular}}} & \multirow{-2}{*}{\cellcolor[HTML]{FFFFFF}\textbf{Model Version}} & \textbf{mean}               & \textbf{std$^*$}               & \textbf{mean}               & \textbf{std$^*$}               & \textbf{mean}                  & \textbf{std$^*$}    & \multirow{-2}{*}{\cellcolor[HTML]{FFFFFF}\textbf{prod$^*{}^*$}}             \\
\hline
\cellcolor[HTML]{FFFFFF}                                                                                   & LibPostal                                                           & 0.997                       & -                           & 0.997                      & -                           & 0.992                         & -   & -                           \\
\cellcolor[HTML]{FFFFFF}                                                                                   & DeepParse                                                        & 0.991                       & -                           & 0.992                      & -                           & 0.727                         & -      & -                         \\
\multirow{-3}{*}{\cellcolor[HTML]{FFFFFF}\textbf{$V_0$}}                                                      & distilbert-base-uncased                                                        & 0.999                      & 0.050                       & 0.999                      & 0.010                       & 0.885                         & 1.536          & -                 \\
\hline
\cellcolor[HTML]{FFFFFF}                                                                                   & LibPostal                                                           & 0.898                       & -                           & 0.897                      & -                           & 0.918                         & -           & -                    \\

\cellcolor[HTML]{FFFFFF}                                                                                   & DeepParse                                                        & 0.828                       & -                           & 0.826                      & -                           & 0.767                         & -             & -                  \\
\multirow{-3}{*}{\cellcolor[HTML]{FFFFFF}\textbf{$V_1$}}                                                      & distilbert-base-uncased                                                        & 0.995                       & 0.425                       & 0.995                      & 0.025                       & 0.924                         & 0.400               & -            \\
\hline
\cellcolor[HTML]{FFFFFF}                                                                                   & LibPostal                                                           & 0.761                       & -                           & 0.759                      & -                           & 0.781                         & -           & 0.675                   \\
\cellcolor[HTML]{FFFFFF}                                                                                   & DeepParse                                                        & 0.747                       & -                           & 0.747                      & -                           & 0.709                         & -             & -                  \\
\cellcolor[HTML]{FFFFFF}                                                                                   & \textbf{distilbert-base-uncased}                                                        & 0.994                       & 0.371                       & 0.994                      & 0.045                       & \textbf{0.859}                         & 0.915            & 0.765               \\
\cellcolor[HTML]{FFFFFF}                                        & distilbert-multi                                        & 0.992           & 0.212                                   & 0.992                       & 0.068                      & 0.827                          & 1.708        & -                  \\
\cellcolor[HTML]{FFFFFF}                                        & distilroberta                                           & 0.992           & 0.188                                   & 0.991                       & 0.054                      & 0.835                          & 2.998           & -               \\
\cellcolor[HTML]{FFFFFF}                                        & distilbert-cased                                        & 0.991                & 1.361                              & 0.990                          & 1.230                   & 0.821   & -          \\
\cline{2-9}
\cellcolor[HTML]{FFFFFF}                                                                                   & distilbert-no-hardsep                                        & 0.994               & 0.338                              & 0.994                       & 0.066                   & 0.860                            & 1.075              & -              \\
\cellcolor[HTML]{FFFFFF}                                                                                   & distilbert-noprefix                                      & 0.995                & 0.401                              & 0.994                          & 0.017                   & 0.862     & 0.938       & -              \\
\cellcolor[HTML]{FFFFFF}                                                                                   & \textbf{distilbert-early-stopping}                                        & 0.976                       & 2.711                       & 0.975                      & 2.649                       & \textbf{0.871}                         & 0.991              & 0.766             \\
\cellcolor[HTML]{FFFFFF}                                                                                   & \multicolumn{1}{c|}{\cellcolor[HTML]{FFFFFF}\begin{tabular}[c]{@{}c@{}}distilbert-no-prefix+\\ early-stopping\end{tabular}}                          & 0.931                       & 11.240                      & 0.929                      & 10.980                      & 0.863                         & 3.027       & -                   \\

\cline{2-9}
\cellcolor[HTML]{FFFFFF}                                                                                   & bert-base-uncased                                                & 0.967                       & 15.768                      & 0.966                      & 15.843                      & 0.868                         & 5.170         & 0.784                 \\
\cellcolor[HTML]{FFFFFF}                                                                                   & bert-base-large                                                  & 0.966                       & 5.013                       & 0.965                      & 4.977                       & 0.909                         & 2.140            & 0.785              \\
\multirow{-10}{*}{\cellcolor[HTML]{FFFFFF}\textbf{$V_2$}}                                                     & \textbf{xlm-roberta-large}                                                & 0.975                       & 4.817                       & 0.974                      & 4.996                       & \textbf{0.924}                         & 2.236     & \textbf{0.801}                     
\end{tabular}
\captionsetup{justification=centering}
\caption*{$^*$~Reported values of standard deviation are multiplied by $10^3$ for clarity  \\
$^*{}^*$~Actual real-life payment address data, referred as \textbf{production} data}
\caption{F1 scores of baseline and Transformer models}
\label{tab: results table}
\end{table*}

\begin{table}
\centering
\begin{tabular}{lc}
\hline
\textbf{Model} & \textbf{zero-shot}\\
\hline
Llama2 7B & 0.4650 \\
Mistral 7B Instruct v0.2 & 0.6066  \\
Mistral 7B Instruct v0.2 (SFT*) & 0.7113  \\
Mixtral 8x7B Instruct v0.1 & 0.7233  \\
\hline
\end{tabular}
\captionsetup{justification=centering}
\caption{F1 score on Decoder Based approach\\ $^*$Fine-Tuned on 1,000 samples with 3 epochs}
\label{tab:plain_llm}
\end{table}

\textbf{- Generative LLMs:}  In Table~\ref{tab:plain_llm} we show the main results of the Generative LLMs on 10,000 samples from dataset $V_2$. As stated earlier Mistral 7-B demonstrated a relatively good performance just with simple prompt engineering. 

Given the repetitive nature of our prompts, we found that the LORA adapters are very susceptible to over-fitting, and can deliver a significant performance improvement with just 1,000 training samples for 3 epochs. Furthermore, we tried several ranks for our adapters and settled for a rank of 8 with the standard LORA value of 2 * r.

One interesting observation is the impact of the inference parameters on the performance and the quality of the output. For the base Mistral-7B model, the performance remains stable with the different parameters with an F1 of approximately 0.6. However, for our fine-tuned model with a LORA adapter, we see the performance changing from an F1 score of 0.67 to 0.71. A full breakdown of the performance with the different sampling configurations is available in the appendix. Finally, for reference, we did a benchmark on the Mixtral-8X7B model with no fine-tuning and we were able to achieve a similar F1 score of 0.72

%Generally, during our experiments, we observed that a larger training dataset did not make any clear difference and this applies even when exploring various rank and alpha LORA parameters. Indeed when using a large sample with low redundancy, the model over-fits. (show snapshot numbers?). We can also see that a trained Mistral 7b gives close results compared to Mixtral 8x7B. We attempt to train Mixtral 8x7B however it performs poorly compared to the shared results.%

As expected during our manual checks we noticed hallucinations by the LLMs. Sometimes the generated text modifies the word's index. In case the word appears only once, we can retrieve its position in the input, however, in case there is an ambiguity the prediction fails. The model may also not follow the instructions by generating an unexpected output format, e.g. a nested JSON. In this case, we attempt to flatten the JSON and retrieve the expected tag. This mostly happens when there is a country code and the full country name in the input address. The output may unexpectedly also include a "Hardsep" "\$" and may create new tags that are not in the input. For example, inferring the country name from the city.

\section{Conclusion}

In summary in this paper, we introduced a new open-sourced dataset reflecting the limitations and noise of real-world payment data. This will enable better benchmarking of expected production performance and further research on this problem. 
%Furthermore, with the use of our augmentation framework, we provide a resource to create other domain-specific datasets.
Our experimental results highlight the importance of training robust models that are capable of dealing with noise. We achieve state-of-the-art performance on the synthetic zero-shot data and our production data with a fine-tuned XLM-RoBERTa-Large model. A derivative of this model is currently deployed in our production systems. \\
Overall even though the generative LLMs were not able to match the performance of the encoder transformer models, we believe there is a strong potential that warrants further inquiry. Possible directions for further improvement would be the use of context-free grammars such as the LMQL library \citep{Beurer_Kellner_2023} to force a more structured output, and the introduction of geography knowledge with a LlamaIndex.\\
%With the final version of this paper, we plan to open-source all training and evaluation code as well as the fine-tuned models.
%we provide a valuable resource for other companies and researchers facing similar challenges in Address Parsing for a wide range of real-world applications besides AML transaction monitoring.
In the final version of this paper, we intend to open-source all training and evaluation code, along with the fine-tuned models. By doing so, we aim to offer a valuable resource to other companies and researchers confronting analogous challenges in Address Parsing, applicable across various real-world applications beyond the financial industry.

\bibliography{anthology,custom}

\let\cleardoublepage\clearpage

\appendix

\section{Appendix}
\label{sec:appendix}
\subsection{Experimental setup}
\label{sec:setup}
We perform a 4-fold cross-validation on each model, where for each fold we tokenize the training and validation sets with the model-specific pretrained tokenizer. We also align the tags in a way that each tag corresponds to the starting token of the word (or the whole word if it was not split), the whole process is shown in Table~\ref{tab:tokenization example}. We use the final format as input to the model, with the hyperparameters depicted in Table~\ref{tab:training_hyperparameters}.

\subsection{Data \& Source Code}
The data and source code can openly be accessed on \href{https://arxiv.org/abs/2404.05632}{https://arxiv.org/abs/2404.05632}.

\begin{table*}[]
\centering
\begin{tabular}{
>{\columncolor[HTML]{FFFFFF}}c 
>{\columncolor[HTML]{FFFFFF}}c 
>{\columncolor[HTML]{FFFFFF}}r 
>{\columncolor[HTML]{FFFFFF}}r }
\hline
\multicolumn{1}{c}{\cellcolor[HTML]{FFFFFF}{\color[HTML]{000000} \textbf{Tag}}}                          & {\color[HTML]{000000} \textbf{Word}}                                                       & \multicolumn{1}{l}{\cellcolor[HTML]{FFFFFF}{\color[HTML]{000000} \textbf{Subword}}} & \multicolumn{1}{r}{\cellcolor[HTML]{FFFFFF}{\color[HTML]{000000} \textbf{Final Tag}}} \\ \hline
\multicolumn{1}{l}{\cellcolor[HTML]{FFFFFF}{\color[HTML]{000000} }}                             & {\color[HTML]{000000} }                                                           & {\color[HTML]{000000} {[}CLS{]}}                                            & {\color[HTML]{000000} UNK}                                                    \\ \hline
\multicolumn{1}{l}{\cellcolor[HTML]{FFFFFF}{\color[HTML]{000000} }}                             & \cellcolor[HTML]{FFFFFF}{\color[HTML]{000000} }                                   & {\color[HTML]{000000} ki}                                                   & {\color[HTML]{000000} B-StreetName}                                           \\ \cline{3-4} 
\multicolumn{1}{l}{\cellcolor[HTML]{FFFFFF}{\color[HTML]{000000} }}                             & \cellcolor[HTML]{FFFFFF}{\color[HTML]{000000} }                                   & {\color[HTML]{000000} \#\#rch}                                              & {\color[HTML]{000000} UNK}                                                    \\ \cline{3-4} 
\multicolumn{1}{l}{\cellcolor[HTML]{FFFFFF}{\color[HTML]{000000} }}                             & \cellcolor[HTML]{FFFFFF}{\color[HTML]{000000} }                                   & {\color[HTML]{000000} \#\#ens}                                              & {\color[HTML]{000000} UNK}                                                    \\ \cline{3-4} 
\multicolumn{1}{l}{\multirow{-4}{*}{\cellcolor[HTML]{FFFFFF}{\color[HTML]{000000} StreetName}}} & \multirow{-4}{*}{\cellcolor[HTML]{FFFFFF}{\color[HTML]{000000} kirchenstr}}       & {\color[HTML]{000000} \#\#tr}                                               & {\color[HTML]{000000} UNK}                                                    \\ \hline
{\color[HTML]{000000} StreetNumber}                                                               & {\color[HTML]{000000} 24}                                                         & {\color[HTML]{000000} 24}                                                   & {\color[HTML]{000000} B-StreetNumber}                                         \\ \hline
\cellcolor[HTML]{FFFFFF}{\color[HTML]{000000} }                                                   & \cellcolor[HTML]{FFFFFF}{\color[HTML]{000000} }                                   & {\color[HTML]{000000} 36}                                                   & {\color[HTML]{000000} B-PostalCode}                                           \\ \cline{3-4} 
\multirow{-2}{*}{\cellcolor[HTML]{FFFFFF}{\color[HTML]{000000} PostalCode}}                       & \multirow{-2}{*}{\cellcolor[HTML]{FFFFFF}{\color[HTML]{000000} 3660}}             & {\color[HTML]{000000} \#\#60}                                               & {\color[HTML]{000000} UNK}                                                    \\ \hline
\cellcolor[HTML]{FFFFFF}{\color[HTML]{000000} }                                                   & \cellcolor[HTML]{FFFFFF}{\color[HTML]{000000} }                                   & {\color[HTML]{000000} gem}                                                  & {\color[HTML]{000000} B-Municipality}                                         \\ \cline{3-4} 
\cellcolor[HTML]{FFFFFF}{\color[HTML]{000000} }                                                   & \cellcolor[HTML]{FFFFFF}{\color[HTML]{000000} }                                   & {\color[HTML]{000000} \#\#ein}                                              & {\color[HTML]{000000} UNK}                                                    \\ \cline{3-4} 
\multirow{-3}{*}{\cellcolor[HTML]{FFFFFF}{\color[HTML]{000000} Municipality}}                     & \multirow{-3}{*}{\cellcolor[HTML]{FFFFFF}{\color[HTML]{000000} gemeinde}}         & {\color[HTML]{000000} \#\#de}                                               & {\color[HTML]{000000} UNK}                                                    \\ \hline
{\color[HTML]{000000} Municipality}                                                               & {\color[HTML]{000000} klein}                                                      & {\color[HTML]{000000} klein}                                                & {\color[HTML]{000000} I-Municipality}                                         \\ \hline
\cellcolor[HTML]{FFFFFF}{\color[HTML]{000000} }                                                   & \cellcolor[HTML]{FFFFFF}{\color[HTML]{000000} }                                   & {\color[HTML]{000000} po}                                                   & {\color[HTML]{000000} I-Municipality}                                         \\ \cline{3-4} 
\cellcolor[HTML]{FFFFFF}{\color[HTML]{000000} }                                                   & \cellcolor[HTML]{FFFFFF}{\color[HTML]{000000} }                                   & {\color[HTML]{000000} \#\#ch}                                               & {\color[HTML]{000000} UNK}                                                    \\ \cline{3-4} 
\cellcolor[HTML]{FFFFFF}{\color[HTML]{000000} }                                                   & \cellcolor[HTML]{FFFFFF}{\color[HTML]{000000} }                                   & {\color[HTML]{000000} \#\#lar}                                              & {\color[HTML]{000000} UNK}                                                    \\ \cline{3-4} 
\multirow{-4}{*}{\cellcolor[HTML]{FFFFFF}{\color[HTML]{000000} Municipality}}                     & \multirow{-4}{*}{\cellcolor[HTML]{FFFFFF}{\color[HTML]{000000} pochlarn}}         & {\color[HTML]{000000} \#\#n}                                                & {\color[HTML]{000000} UNK}                                                    \\ \hline
\cellcolor[HTML]{FFFFFF}{\color[HTML]{000000} }                                                   & \cellcolor[HTML]{FFFFFF}{\color[HTML]{000000} }                                   & {\color[HTML]{000000} ni}                                                   & {\color[HTML]{000000} B-Province}                                             \\ \cline{3-4} 
\cellcolor[HTML]{FFFFFF}{\color[HTML]{000000} }                                                   & \cellcolor[HTML]{FFFFFF}{\color[HTML]{000000} }                                   & {\color[HTML]{000000} \#\#ede}                                              & {\color[HTML]{000000} UNK}                                                    \\ \cline{3-4} 
\cellcolor[HTML]{FFFFFF}{\color[HTML]{000000} }                                                   & \cellcolor[HTML]{FFFFFF}{\color[HTML]{000000} }                                   & {\color[HTML]{000000} \#\#ros}                                              & {\color[HTML]{000000} UNK}                                                    \\ \cline{3-4} 
\cellcolor[HTML]{FFFFFF}{\color[HTML]{000000} }                                                   & \cellcolor[HTML]{FFFFFF}{\color[HTML]{000000} }                                   & {\color[HTML]{000000} \#\#ter}                                              & {\color[HTML]{000000} UNK}                                                    \\ \cline{3-4} 
\cellcolor[HTML]{FFFFFF}{\color[HTML]{000000} }                                                   & \cellcolor[HTML]{FFFFFF}{\color[HTML]{000000} }                                   & {\color[HTML]{000000} \#\#re}                                               & {\color[HTML]{000000} UNK}                                                    \\ \cline{3-4} 
\multirow{-6}{*}{\cellcolor[HTML]{FFFFFF}{\color[HTML]{000000} Province}}                         & \multirow{-6}{*}{\cellcolor[HTML]{FFFFFF}{\color[HTML]{000000} niederosterreich}} & {\color[HTML]{000000} \#\#ich}                                              & {\color[HTML]{000000} UNK}                                                    \\ \hline
\multicolumn{1}{l}{\cellcolor[HTML]{FFFFFF}{\color[HTML]{000000} }}                             & {\color[HTML]{000000} }                                                           & {\color[HTML]{000000} {[}SEP{]}}                                            & {\color[HTML]{000000} UNK}                                                    \\ \hline
\end{tabular}
\caption{Tokenization example}
\label{tab:tokenization example}
\end{table*}

\begin{table*}[ht]
\centering
\begin{tabular}{
>{\centering\arraybackslash}p{4.5cm}
>{\centering\arraybackslash}p{4.5cm}
>{\centering\arraybackslash}p{4.5cm}
}
\hline
\textbf{\cellcolor[HTML]{FFFFFF} Postal tags} & \textbf{\cellcolor[HTML]{FFFFFF} $V_0$ and $V_1$ mapping} & \textbf{\cellcolor[HTML]{FFFFFF} $V_2$ mapping} \\
\hline
house\_number & StreetNumber & StreetNumber \\
road & StreetName & StreetName \\
house & Unit & Name \\
level & Unit & Unit \\
city & Municipality & Municipality \\
state & Province & Province \\
state\_district & Province & Province \\
unit & Unit & Unit \\
postcode & PostalCode & PostalCode \\
country & Province & Country \\
suburb & Municipality & Municipality \\
city\_district & Municipality & Municipality \\
category & StreetName & OOA \\
near & Municipality & OOA \\
po\_box & PostalCode & OOA \\
entrance & Unit & OOA \\
country\_region & Province & Country \\
staircase & Unit & OOA \\
world\_region & Province & Province \\
island & Province & OOA \\
\hline
\end{tabular}
\caption{Postal tags mapping}
\label{tab:postal-mapping}
\end{table*}

\begin{table*}[ht]
    \centering
    \begin{tabular}{cc}
        \hline
        \textbf{Hyperparameter} & \textbf{Value} \\
        \hline
        num train epochs & 1 \\
        train batch size & 1024 \\
        evaluation batch size & 1024 \\
        dropout & 0.1 \\
        learning rate scheduler warmup steps & 500 \\
        optimizer & adamw \\
        optimizer weight decay & 0.01 \\
        evaluation steps & 20 \\
        early stopping patience$^*$ & 5 \\
        seed & 42 \\
        \hline
    \end{tabular}
    \caption*{$^*$~only when early stopping is stated to be used}
    \caption{Training Hyperparameters}
    \label{tab:training_hyperparameters}
\end{table*}

\begin{figure*}[t]
\begin{lstlisting}
f"""<s>[INST]
    You are a word classifier that classifies words from a text corresponding to an address free text field.
    You should analyze with deep precision the INPUT and return a dictionary with the following keys: "Name", "StreetNumber", "StreetName", "Municipality", "PostalCode", "Unit", "Country", "CountryCode".
    Each word is separated by a space and should be classified without any modification.
    Each word in the input has a prefix with the index i as '[i]-' and it should be ignored for the classification but it should remain AS-IS in the output.
    Sub sequence of words should be classified as follow:
    'Name': words corresponding to an indiviual name or institution name.
    'StreetNumber': words corresponding to a street number.
    'StreetName': words corresponding to a street name.
    'Municipality': words corresponding to a municipality or city.
    'PostalCode': words corresponding to a postal code.
    'Unit': words corresponding to a unit number.
    'Country': words corresponding to a full country name.
    'CountryCode': words corresponding to a country iso2 code.
    
    Output Indicator:
    2. Usually a name comes before the address.
    3. "$" is indicating a large separator and it should not be classified.
    4. The output words should be taken from the input only and it should not be modified
    5. The same word cannot be used in two different classes.
    6. Words are classified subsequently.
    7. Empty classes should not appear in the output.
    8. Output should not include nested values.
    9. Each index are taken from the input itself and the index matches, e.g. the prefix '[i]-' remains unchanged for all words.
    
    For example:
    ### INPUT:
    "[0]-THOMASSEN [1]-GULBRANDSEN [2]-OG [3]-GUNDERSEN [4]-$ [5]-TV [6]-SD [7]-9 [8]-JAPARATINGA [9]-57950 [10]-000 [11]-BR"
    ### OUTPUT: 
    {{"Name": "[0]-THOMASSEN [1]-GULBRANDSEN [2]-OG [3]-GUNDERSEN", "StreetName": "[5]-TV [6]-SD [7]-9", "Municipality": "[8]-JAPARATINGA", "PostalCode": "[9]-57950 [10]-000", "CountryCode": "[11]-BR"}}
    
    
    ### INPUT:
    {address}
    [/INST]
    
    ### OUTPUT:
    """
\end{lstlisting}
\caption{Prompt Template used for LLMs}
\label{fig:prompt}
\end{figure*}

\begin{table*}[t]
\centering
\begin{tabular}{lll||lll}
\textbf{Model}                  & \textbf{min\_p$^\dagger$} & \textbf{top\_p$^\ddagger$} & \multicolumn{3}{c}{\textbf{Temperature}} \\
                                &                 &                 & {0.8}        & {0.5}       & {0.2}       \\ \hline \hline
Llama-2-7b-chat-hf              & 0.1             & -               & -            & -           & 0.465       \\
Llama-2-7b-hf                   & 0.1             & -               & -            & -           & 0.424       \\ \hline
Mistral-7B-Instruct-v0.2        & 0.1             & -               & 0.601       & 0.601      & 0.607      \\
Mistral-7B-Instruct-v0.2        & 0.3             & -               & 0.605       & 0.601      & 0.602      \\
Mistral-7B-Instruct-v0.2        & 0.5             & -               & 0.601       & 0.606      & 0.602      \\
Mistral-7B-Instruct-v0.2        & -               & 0.9             & 0.603       & 0.603      & 0.604      \\
Mistral-7B-Instruct-v0.2        & -               & 0.7             & 0.601       & 0.606      & 0.604      \\
Mistral-7B-Instruct-v0.2        & -               & 0.5             & 0.605       & 0.605      & 0.604      \\ \hline
Mistral-7B-Instruct-v0.2 (SFT*) & 0.1             & -               & 0.675       & 0.696      & 0.708      \\
Mistral-7B-Instruct-v0.2 (SFT*) & 0.3             & -               & 0.698       & 0.702      & 0.709      \\
Mistral-7B-Instruct-v0.2 (SFT*) & 0.5             & -               & 0.702       & 0.706      & 0.710      \\
Mistral-7B-Instruct-v0.2 (SFT*) & -               & 0.9             & 0.671       & 0.695      & 0.709      \\
Mistral-7B-Instruct-v0.2 (SFT*) & -               & 0.7             & 0.700       & 0.710      & 0.709      \\
Mistral-7B-Instruct-v0.2 (SFT*) & -               & 0.5             & 0.710       & 0.709      & 0.708      \\
\end{tabular}
\captionsetup{justification=centering}
\caption{F1 score of LLMs with various text generation parameters. Variants of Mistral 7B without SFT is not significant while the Fine-Tuned versions are more sensible to text generation parameters change.\\ $^*$ Trained on 1,000 samples and 3 epochs \\ $^\dagger$ Minimum probability for a token to be considered, relative to the probability of the most likely token. top\_p does not vary when min\_p is set. \\ $^\ddagger$ Cumulative probability of top tokens to be considered. min\_p does not vary when top\_p is set.}
\label{tab:llm_all_results}
\end{table*}

\end{document}